# Morphosyntactic Analysis for CHILDES

Houjun Liu
Stanford University

Brian MacWhinney
Carngie Mellon University

**Abstract:** Language development researchers are interested in comparing the process of language learning across languages. Unfortunately, it has been difficult to construct a consistent quantitative framework for such comparisons. However, recent advances in AI (Artificial Intelligence) and ML (Machine Learning) are providing new methods for ASR (automatic speech recognition) and NLP (natural language processing) that can be brought to bear on this problem. Using the Batchalign2 program (Liu et al., 2023), we have been transcribing and linking data for the CHILDES database and have applied the UD (Universal Dependencies) framework to provide a consistent and comparable morphosyntactic analysis for 27 languages. These new resources open possibilities for deeper crosslinguistic study of language learning.

**Corresponding author:** Brian MacWhinney, Department of Psychology, Carnegie Mellon University, Pittsburgh, PA. Email: macw@cmu.edu

**ORCID ID:** https://orcid.org/0000-0002-4988-1342

# Introduction

Child language research involves three, partially separate, formats for data collection. The first focuses on the development of a single child or pair of children, often across several years. Work in this tradition includes classic diary studies from German (Stern & Stern, 1907), French (Bloch, 1921; Guillaume, 1927), Polish (Smoczynska, 2017; Szuman, 1959), Hungarian (Kenyeres, 1926; Ponori, 1871), Mandarin (Chao, 1951), Bulgarian (Gvozdev, 1949), Serbian (Pavlovitch, 1920) and other languages. It also includes diary and transcript studies of particular aspects of development such as phonology (Smith, 1973), grammatical morphology (Brown, 1973), lexicon (Tomasello, 1992), or all of the above (Leopold, 1939, 1947, 1949a, 1949b). This case-study work has helped us understand the diverse ways in which children acquire and use language to express their needs (Karniol, 2010).

A second data collection format measures and evaluates learning across groups of children within a single language. This type of analysis is particularly important for clinicians who need to diagnose, assess, and remediate language learning disorders. Data collection in this format includes standardized tests (Bishop, 1982; Goldman & Fristoe, 2000), language sample analysis (Garbarino et al., 2020), and language profiling (Bernstein Ratner & MacWhinney, 2023; Crystal et al., 1989; Scarborough, 1990).

A third data collection format examines development across languages. This work considers the ways in which variations in language structure and social input pose challenges or opportunities to the learner. For various reasons, this work has had a concentration of data from WEIRD (Western, educated, industrialized, rich, and democratic) participants (Henrich et al., 2010) along with an emphasis on monolingual acquisition. To broaden our crosslinguistic coverage, Slobin and colleagues (Slobin, 1985) have provided descriptions of linguistic and social development in a series of languages, including some from non-WEIRD communities. However, without quantitative tools to compare across these many languages, it has been difficult to generalize about patterns of language learning methods, structures, and challenges. The introduction of the MacArthur-Bates Communicative Development Inventory (Dale & Fenson, 1996) provided quantitative methods to bridge the WEIRD gap for the earliest stages of lexical development. That tool has now been validated for several Western languages (Frank et al., 2021), but extensions to less well-resourced languages and multilingualism (Tamis-LeMonda et al., 2024) will take additional time and effort.

The CHILDES data-sharing system (MacWhinney, 2000) offers yet another approach to extending child language research beyond WEIRD participants. CHILDES includes language samples from 49 languages, along with 41 corpora from children learning two or more languages, all contributed by researchers who are speakers of these languages. Although many of these families are WEIRD, there are also many from societies that are not Western, and not fully industrialized, rich, or democratic. Although nearly 40% of the data is from English, there are many large corpora from languages such as Mandarin, Spanish, German, French, and Japanese as well as a smaller number of large corpora from another 15 languages.

Creating child language corpora requires major commitments of researcher effort for recording, transcription, and analysis. However, recent advances in AI (artificial intelligence) and ML (machine learning) have led to marked improvements in ASR (automatic speech recognition)(Radford et al., 2022) and NLP (natural language processing)(Nivre et al., 2016) methods that can markedly facilitate this work. The use of ASR can greatly speed transcription (Liu et al., 2023), although recognition of child vocalizations before age 3 is still poor. When recording is done well, ASR can recognize adult input accurately enough to allow a transcript to be finalized after a much briefer period of hand correction. A further advantage is that ASR it creates a transcript that is linked to the audio on both the utterance and single word level, thereby facilitating analyses of phonology, fluency, and total time talking. Moreover, the output can be structured directly in the CHAT (Codes for Human Analysis of Talk) format, thereby allowing analysis through the utilities built into the CLAN (Child Language Analysis) program (MacWhinney & Fromm, 2022). ASR methods can also be used to automatically link an unlinked transcript to the corresponding media (audio or video) on the utterance and word level. This process is particularly useful for transcripts in the CHILDES database that have media, but which have not yet been linked to that media.

After a transcript has been created in correct CHAT format, we can then use NLP methods to automatically construct a complete morphosyntactic analysis. In the next sections, we will describe how these ASR and NLP methods are being applied to improve the use of CHILDES data across all three data analysis formats with a special emphasis on facilitating crosslinguistic comparisons.

### Automatic Speech Recognition

Once a language sample has been recorded, the next task is to create a transcript. Depending on the nature of the interaction, manual transcription of one hour of interaction can take from 10 to 16 hours (Bernstein Ratner & MacWhinney, 2020). To speed up this process, researchers can apply ASR methods using the Batchalign2 system (Liu et al., 2023) which outputs a transcript in the CHAT format required for inclusion in the CHILDES database. Batchalign2 offers access to two ASR systems: the Rev.AI ASR cloud service (Del Rio et al., 2022) or a local ASR model based on OpenAI Whisper (Radford et al., 2023). If IRB (Institutional Review Board) regulations do not allow transmission of data to a cloud service, users may prefer to use Whisper, although Rev.AI explicitly allows the user to determine that the data will not be stored on their cloud server. For English, Rev.AI output is a bit more accurate than Whisper due to the its use of a large amount of two-party conversations as training data (Del Rio et al., 2022). In addition, processing through Rev.AI is much faster than running with Whisper when local hardware is limited, but both options are good choices.

Another factor that favors use of Whisper is that the training data for the NLP models used in downstream analysis use native orthographies of each language (De Marneffe et al., 2021; Qi et al., 2020). Latinized transcripts must be converted back into the standard orthography for the language before downstream analysis. Because of this limitation, the significantly wider language and orthographic profile of the Whisper model (in particular, WhisperV3 available at https://huggingface.co/openai/whisper-large-v3) is advantageous for non-English languages; therefore, the majority of the recognition needed to cover all the languages described here (and in particular ones with non-latinized native orthography) is performed with the Whisper option.

**Utterance Segmentation**
Tagging for morphological categories and grammatical dependency structure requires accurate delineation of sentences or utterances. Segmentation of naturalistic spoken language data requires attention to features not found in written text (Fraser et al., 2015), such as incompletion, repetition, retracing, and other features. Sections 9.1 and 9.2 of the CHAT manual (https://talkbank.org/manuals/CHAT.pdf) provide a set of standards for utterance segmentation. For example, one important feature is that clauses joined only with coordinating conjunctions (and, or, but) are treated as separate *utterances*.

Because currently available tokenizers are all based on written language and because spoken language segmentation follows quite different rules and patterns, we have created novel tokenizers based on spoken language training data. To create the tokenizer for spoken English data, we turned to the TalkBank database, which contains many Gold Standard utterances segmented according to the rules mentioned above. The tokenizer (Liu et al., 2023) is trained via a token-classification task, which assigns each input text token as being the start (label 1), middle (label 0), a phrase which should be separated by a comma (label 5), or end of each utterance (label 2,3,4); in particular, there are three utterance-ending labels, each corresponding to the utterance being declarative, interrogative, or exclamatory respectively. The tokenizer uses a BERT-class model (Devlin et al., 2018) to generate semantic embeddings for language modeling, and a deep neural network (DNN) to perform token-level annotations.

Currently, Batchalign2 provides tokenizers for English and Mandarin. The English model was trained on the MICASE (The Michigan Corpus of Academic Spoken English) (Römer, 2019) corpus in CABank (https://ca.talkbank.org/access/MICASE.html), which includes transcribed data from 300 participants in a wide variety of interactions between students and faculty at the University of Michigan. The Mandarin model was trained on three corpora available on the TalkBank CHILDES database—Zhou Assessment (Li & Zhou, 2011), Chang Personal Narrative (Chang & McCabe, 2013), and Li Shared Reading. The ability to train new segmentation models based on segmented CHAT transcripts has been released along with the Batchalign2 software. In addition, work currently in progress by the HuggingFace diarization team (https://github.com/huggingface/diarizers) using the pyannote framework (Bredin, 2023) with TalkBank data should be able to provide tokenizers for a wider variety of languages.

### Text-Media Alignment

Apart from the processing of new recordings, ASR can be useful for linking previously hand-transcribed transcripts to media for timing-aware analysis. The Batchalign2 "align" command supports this process by running a two-pass alignment of transcripts to media. The first pass of this process involves performing rough, utterance *time* diarizations using ASR as a silver annotation reference. The second pass involves latent feature extraction of each form's timestamp within the utterance by using latent attention activations from the Whisper ASR model described above.

**Utterance Timing Recovery**

We begin by assuming that the transcript to be linked has correctly segmented utterance text, but that it does not yet have any utterance time values. If the transcript has imprecise time values, we can use the CLAN CHSTRING command with the +cbullets.cut switch to remove them. We must then identify the relative time within the media in which an utterance occurred. This task is difficult to perform with classic alignment schemes, which face difficulty generating correct alignments among longer timestamps without some form of hierarchical or recursive scheme (Moreno et al., 1998), due to the exponential growth in number of possible alignments as sequence length increases.

To address this limitation, we take an optimistic, silver-labeling approach by using an ASR-generated transcript (which can process the audio linearly by splitting it into segments) to obtain a silver transcript which we call the "backplate." Because this ASR transcript has been generated directly from the audio, each of its tokens are linked against a relative timestamp within the audio file. By then aligning the transcript against the backplate, we can induce the timestamp in which each utterance in the gold standard transcript exists by reading the corresponding times on the backplate.

To perform the actual transcript-to-transcript alignment described above, we apply dynamic programming (Bellman, 1966) to create an alignment solution which minimizes the form-level Levenshtein edit distance (Jurafsky & Martin, 2009) between the gold transcript and the backplate. We can then calculate the level timings via direct computation using the first and last timestamps of aligned forms within an utterance labelled by the gold transcript, plus some time on each end to account for errors which will be tightened in the second step of the overall alignment procedure.

Although this procedure could theoretically also recover the timing of each individual token (simply by aligning the backplate transcript against gold at a token level) this initial alignment is only practically feasible for utterance timing recovery. Instead, we assume that the overall time alignment for an utterance (as denoted by the timing between its first aligned token and the last aligned token) should be roughly accurate. Because we are doing utterance level alignment, any errors in the backplate (such as missing a filled pause, a very common error in ASR) which are within the bounds of an utterance are essentially irrelevant to this procedure. Even if a particular utterance is not properly transcribed in the backplate, we can infer its temporal alignment by knowing the values for the previous and following utterances. In comparison, application of this procedure on the token level would result in missing time values for all forms which do not have precise alignments between the gold and backplate transcripts—reducing the quality of the resulting data.

### Word-level Forced Alignment
Next, to obtain word-level or token-level alignment, we use latent attention analysis through the Whisper ASR model. Recall that Whisper is an encoder-decoder architecture model (Radford et al., 2023), whereby the encoder creates a latent embedding per sample (usually 16,000Hz) of the input audio sequence which is then used as input to the cross-attention (Niu et al., 2021) computation against the output text sequence.

The key motivation of our analysis follows closely to previous work in cross-attention activation analyses (Hou et al., 2019). We take advantage of the heuristic that the *most highly activated (high value) encoder-decoder cross-attention pairs are likely the most directly relevant pairings*. In the context of speech analysis, this means that the most highly activated encoder time slice to decoder token activation is likely the best temporal alignment for the token.

To take advantage of this fact, we run a single forward pass on the Whisper model per time-segmented utterance, supplying the utterance time slice (derived in the previous step of the overall alignment procedure) as the encoder input and the gold utterance text as the decoder input. Then, we extract the last cross-attention activation matrix from the model activations during this forward pass.

From this, we apply a series of normalization procedures — mean centering and median filter smoothing (Brownrigg, 1984) — to obtain a smoothed cross-attention matrix. Taking highest values indices of this matrix along each axis reveals two sequences — one for time along each slice and another for transcript-token along each slice. Finally, alignment between these two sequences — which are already sorted in temporal order with alignments between them given by the matrix — will provide a resolved time-per-token value given by Dynamic Time Warping (DTW).

This procedure is relatively quick to compute. Although DTW has O(nm) time complexity, the sequences are reasonably short, and they do not require perfect ASR performance because the gold transcript is provided directly to the Whisper decoder. Through this scheme, we obtain a precise time alignment for each input form which can be used in downstream analysis.

### Universal Dependencies

Next, we will explain how Batchalign2 operates to produce morphosyntactic analyses. This work relies on the application of Universal Dependency (UD) models trained through the Stanza Python NLP package (Qi et al., 2020). This system, which can be used with over 70 languages (https://universaldependencies.org), is based on a consistent language-general set of codes for POS (parts of speech), GFs (grammatical features), and GRs (grammatical relations). Stanza models for each UD language can be downloaded for use by the Batchalign2 Python program which is freely available for download from https://github.com/talkbank.  Before reviewing the details of the application of UD tagging to CHILDES data, we need to consider the previous state-of-the-art for tagging CHILDES transcripts.

Beginning in 1995, Brian MacWhinney, Roland Hausser, and Mitzi Morris created a system for word-level morphological coding called MOR (MacWhinney, 2008). This system relied on a series of hand-crafted declarative rules governing possible word analyses and a program called POST, created by Christophe Parisse (Parisse & Le Normand, 2000) for disambiguating alternative readings in context. The resultant analyses were entered on a %mor line in which each word on the main speech line is given its own morphological analysis. The manual for MOR is available at https://talkbank.org/manuals/MOR.pdf. Across the years, Leonid Spektor extended the the MOR program and Brian MacWhinney refined the lexicon and rules to achieve a high level of accuracy and coverage.  However, extending MOR to other languages represented a major challenge.  Versions of MOR were created for French, Hebrew, Italian, Japanese, and Mandarin. However, these additional versions of MOR were created by a single person and learning how to build a new MOR grammar is difficult. Given this, extensions to the remaining 44 languages in CHILDES are outside the current scope of the project.

The creation of automatic programs for syntactic analysis across these 49 languages faced similar hurdles. Sagae and colleagues (Sagae et al., 2010) created a program called MEGRASP (maximum entropy grammatical relations syntactic processor) that uses the SVM (Support Vector Machine) method to tag CHILDES English and Spanish corpora for grammatical relation dependency structure. In principle, MEGRASP could be extended to cover additional languages.  However, settling on consistent labels for the grammatical relations in each language and applying those labels to a large corpus of training utterances represented yet another major task that would have to be done one-by-one for all the languages in CHILDES.

Given the scope of the work needed to build MOR and MEGRASP analyzers for 49 languages and for languages that will be added to CHILDES in the future, we looked for alternative methods for building morphosyntactic analyses across languages. Fortunately, the UD Project provides almost exactly what was needed. Relying on the latest AI/NLP technology, the UD community has been working to create taggers for 70 languages, including a majority that are outside of Indo-European. UD uses six open class POS (part-of-speech) tags (ADJ, ADV, INTJ, NOUN, PROPN, and VERB) and eight closed class POS tags (ADP, AUX, CCONJ, DET, NUM, PART, PRON, and SCONJ. It clusters GFs into seven lexical feature sets (PronType, NumType, Poss, Reflex, Foreign, Abbr, and Typo), nine nominal inflectional feature sets (Gender, Animacy, NounClass, Number, Case, Definite, Deixis, DeixisRef, and Degree) and ten verbal inflectional feature sets (VerbForm, Mood, Tense, Aspect, Voice, Evident, Polarity, Person, Polite, and Clusivity). Within each set, a further set of GF values is described. For example, Gender has the values Masc, Fem, Neut, and Com. Apart from this systematic listing of POS and GFs, UD provides a uniform nomenclature for grammatical relations (GRs) with six core arguments (nsubj, obj, iobj, csubj, ccomp, and xcomp), ten non-core dependents (obl, vocative, expl, dislocated, advcl, advmod, discourse, aux, cop, and mark), and ten coordination relations (conj, cc, fixed, flat, list, parataxis, compound, orphan, goeswith, and reparandum). The UD web pages provide complete descriptions of all these POS, GFs, and GRs and the documentation for each language shows how they map onto the language.

**Preparing for UD Analysis**

To align with the various format requirements of UD, Stanza, and Batchalign2, we first require transcripts need to be in full compliance with the CHAT format as validated through the Chatter program which is available for download from https://talk-bank.org/software/chatter.html. Because the CHILDES database had been validated using earlier versions of Chatter that failed to enforce some of these requirements, we had to sharpen the specifications in Chatter and reapply the new version to the entire CHILDES database. That process involved a series of format fixes, such as systematization of spacing, use of new fluency codes, and elimination of use of the plus sign for marking compounds. To permit alignment of text to audio, we also needed to eliminate use of repetition codes such as [x 3] for three repetitions of a word or phrase and make overlap and retracing marking more consistent.

Once the data are in the required format, we can run the "morphotag" command in Batchalign2. Internally, this process creates data in the CONLL-U format which is then reformatted to the CHAT format to be written out in the %mor and %gra lines. The POS and GF information is formatted into the %mor line and the GR information is outputted to the %gra line.

Matching the requirements of the UD grammars with the tokenization and transcripts in the CHILDES files faces problems that vary from language to language. One challenge found in nearly all the corpora is the use of eye-dialect to transcribe spoken forms. For example, in English some corpora may have used an apostrophe to represent conversion of final /ŋ/ to final /n/ as in *singin'* which then had to be converted to *singin(g)*. Or German *hab'n* would be converted to *hab(e)n* for consistent recognition by the UD grammar. A form such as *tactor* could be converted to *t(r)actor,* whereas *practor* would be *practor [: tractor].* For languages such as French, Italian, and Spanish that had already gone through analysis by MOR, these conversions were already done, but for other languages they had to be done from scratch.

For the Romance languages - French, Italian, Catalan, Portuguese, and Spanish - there were often issues relating to clitics and portanteau forms. For example, the French corpora often inserted a space between preclitics and stems, as in *j' ai* rather than *j'ai* with the latter being the form expected in standard French orthography. Such divergences were easy enough to fix using global replacements. More complicated cases involved conversions such as *qu'est-ce-que* into *qu'est que*. In each case, the goal of the conversions was to produce output that would match standard orthography, because this is how UD is trained and what it expects.

Another issue facing UD analysis involved how best to handle multi-word expressions (MWE) which the NLP literature refers to as multi-word tokens (MWT). For example, the French word for today is *aujourd'hui,* but without entering this form specifically as an MWT, Stanza's models would separate the front part as the prepositional phrase *au jour* (on the day) and then was unable to tag the remaining segment *d'hui*. To address this problem, we introduced a modification in the Stanza pipeline that allowed for a specified set of MWTs which is checked against during its analysis for each language before downstream analysis such as lemmas, POS, dependencies, and features which would block this form of over-analysis.

It was also necessary to make sure that the word-level transcription for each language matched the standard orthography used for that language, because each UD grammar was trained on data in the standard orthography. This meant that CHILDES corpora that had been transcribed in a Latin or Roman orthography needed to be converted back to the standard orthography for that language. For some languages, this conversion was simple, but for others it represented a greater problem.

**Current State of UD Tagging**
Here we summarize the status of the conversion and tagging process for the 27 languages in CHILDES that have available UD grammars. The 10 languages that have UD grammars, but which have not yet been processed with UD are identified with asterisks. The other 27 have been either fully or partially tagged. These UD taggings represent first drafts that have not yet been checked by native speakers and which will surely require further fine-tuning and use of the MWT method described above. At this point, no further conversion work will be needed for these 27 languages, and they can all go smoothly through future automatic analysis when new versions of UD have been fine-tuned for each language.

1. Afrikaans: Given its limited morphology and the limited use of eye-dialect in transcription, application of UD to Afrikaans went smoothly.
2. *Arabic: The two current Arabic corpora use a romanization which will have to be converted to Arabic script and analyzed through methods that rely on right-to-left orthography.
3. *Basque: There are no obvious barriers to application of UD to Basque, but guidance from native speakers would make the result more reliable.
4. *Bulgarian: The Bulgarian romanization must be converted back to Cyrillic. Unfortunately, there are conflicting standards for romanization and many digraphs are ambiguous, so this conversion will require further analysis.
5. Cantonese: Because the Cantonese corpora were transcribed in Hanzi, no script conversion was necessary. In addition, UD for Chinese languages handles word-level tokenization directly, so there are no issues about any need to add or remove spaces between words.
6. Catalan: Processing of Catalan was straightforward.
7. Croatian: Processing of Croatian was straightforward.
8. Czech: Processing of Czech was straightforward. However, the contributors of the Czech corpus had already created a carefully done %mor analysis which they preferred to keep in place without the UD tags.
9. Danish: Processing of Danish was straightforward.
10. Dutch: Processing of Dutch was straightforward.
11. English: To maintain backward compatibility of the English corpora, we keep the current %mor and %gra lines and add in the new UD lines as %umor and %ugra. The %mor line provides greater morphological detail than the %umor line, particularly for compounds (which are not analyzed by UD). However, the %ugra line is more accurate than the %gra line. Making the UD lines available is important for facilitating cross-linguistic analyses with the other languages, all of which have UD tagging.
12. Estonian: Processing of Estonian was straightforward.
13. French: The French database is quite extensive. However, after much detailed repair, processing went smoothly.
14. German: The German corpora required extensive revision of eye-dialect forms. Once that was done, processing went smoothly. UD did a much better job than the previous MOR in its assignment of case/number/gender roles to modifiers and nouns, as well as in creating an accurate %gra line.
15. *Greek: Processing of Greek will depend on creation of a method for converting from the romanization back to the Modern Greek alphabet.
16. *Hebrew: Hebrew has already been processed by a MOR grammar. However, UD processing of Hebrew will require conversion from romanization to Hebrew script and we have not yet located a method for doing this.

17. *Hungarian: The current Hungarian transcripts make extensive use of eye-dialect and phonological forms. Once these are modified, processing should be straight-forward.

18. Icelandic: Processing of Icelandic required extensive modification of eye-dialect forms that will need to be re-checked. Otherwise, analysis was straightforward.

19. *Indonesian: The huge size of the Indonesian corpus and the extensive use of eye-dialect will require a fair amount of work for this corpus.

20. Irish: Processing of Irish was straightforward.

21. Italian: Processing of Italian was straightforward. Because Italian had earlier been analyzed by MOR, there were few word level problems, except for dealing with separation of clitics by spaces.

22. Japanese: Processing of Japanese has represented a unique challenge because of the use of three orthographies (Kanzi, hiragana, katakana) and difficulties with word segmentation. Two of the Japanese corpora have been tagged, but others will need further orthographic work.

23. Korean: Korean involved no script transformation and processing went quite smoothly.

24. Mandarin: Because Mandarin had already been processed through MOR, there were few irregularities in the transcripts. Also, Mandarin involved no script trans-formation and processing went quite smoothly.

25. Norwegian: Processing of Norwegian was straightforward.

26. Polish: Processing of Polish was straightforward.

27. Portuguese: After some repair for clitics, MWEs, and format, processing of Portu-guese was straightforward.

28. *Romanian: Processing of Romanian is currently in progress.

29. *Russian: Like Bulgarian, Russian will need conversion of romanization to Cyril-lic. However, the extensive use of eye-dialect and phonological forms in the Rus-sian corpora will make this difficult.

30. Serbian: Serbian UD allows for Roman orthography. As a result, processing of Serbian was straightforward.

31. Slovenian: Processing of Slovenian was straightforward.

32. Spanish: Most of the Spanish corpora had earlier been analyzed by MOR. For those corpora, processing was straightforward. However, there are several Span-ish corpora that will need further work for eye-dialect, phonological forms, and other divergences.

33. Swedish: Processing of the Andren corpus was straightforward. However, work with the Lund corpus will require treatment of eye-dialect and phonological forms.

34. *Tamil: Processing of the Tamil transcripts will require conversion of the roman-ization to Abugida orthography.

35. *Thai: Like many other Asian languages, Thai orthography does not include spac-ing, which makes tokenization difficult. Current Thai transcripts all use romanization and there is no clear path for conversion to Sukhothai script.
36. Turkish: Processing of Turkish was straightforward. However, because UD morphology is non-analytic, the %mor line fails to capture the agglutinative nature of Turkish word formation. A similar problem arises with Hungarian and Estonian.
37. Welsh: Processing of Welsh was straightforward, even though there are many forms that involve apostrophes for omissions. Apparently, these forms are already accepted in standard Welsh in the training set for UD.

## Morphosyntactic Analysis

Here we describe in further detail the application of the neural analysis models provided by the Stanza (Qi et al., 2020) system, along with the modifications we make for characteristics of spoken language, child language, and language-specific forms.

### Word Tokenization

The first step of analysis involves tokenizing each utterance in the CHAT transcript into tokens. Because the CHAT format (https://talkbank.org/manuals/CHAT.pdf) encodes tokenization by using whitespace delineated token groups to identify words, tokenization is frequently given natively in the transcript.

However, for some languages token representations have little to do with word-level representations. In Japanese child language, for instance, two of the language's three writing systems—hiragana and katakana—are moraic-based units frequently employed to transcribe a child during L1 development (Ota, 2015) while the third—kanji, often used for actual word representations needed for morphosyntactic analysis, have little to do with phonology. Moreover, Japanese is not written with spaces. Because of this, whitespace-delineated token representations are not a reliable source of information for word representations.

For languages which have this limitation—and in particular, for our analysis of Japanese—we employ the more complex token segmentation scheme given in Stanza which involves formulating word-level tokenization as a token labeling task—ignoring any transcribed tokenizations and labeling each input *character* as belonging to the start, middle, or end of a token—before further processing each resulting "token group" via the downstream, semantic aware modules such as the Stanza lemmatizer. For instance, consider the Japanese phrase *karuto dantai* "cult group" :
カルト団体
The DNN tagger would first treat all constituent forms as separate and assign to each one a beginning and inside tags representing word boundaries. This creates the sequence:
B I I B I
Finally, separating the forms following the B tags, we obtain:
[カルト] [団体]
as the final word tokenizations, which we place back into the CHAT file as space-delimited tokens as follows:
カルト 団体

In this way, we recover a canonical tokenization for those particular languages based on the annotation style chosen by the working group of the target language in UD annotation; for Japanese, for instance, this may include some resulting orthographic Kanji formed by joining tokens from other syllabaries following the short-unit word (SUW) style (Den et al., 2008). We then use this canonical tokenization to "retokenize" the original CHAT transcript with this new tokenization. Once this initial re-tokenization is obtained, we can then proceed to the remaining analysis by the pipeline describe here.

**Multi-Word Token and Form Correction**
UD (De Marneffe et al., 2021) distinguishes between tokens—continuous character spans without delineation in between—and syntactic words used in analysis. This distinction is particularly relevant with respect to the treatment of multi-word tokens (MWTs)—a single continuous text span which contains multiple syntactic words, each with individual features and dependencies which need to be analyzed independently. Augmenting Stanza's neural-only analysis, we use a lexicon and orthography driven approach to identify and expand three types of such MWTs.

Two types of such MWTs are usually automatically recognized by Stanza through the same tokenization procedure described in the section above: clitics and contractions. Clitics are independent syntactical forms attached to other words, such as in Spanish *despertarme* (*despertar* + *me*)—with the latter being a separate syntactic word which modifies the previous word which needs to be analyzed independently (i.e. modifying that I am who woke the object up); contractions are combinations of multiple words into one token, such as in English *I'm* (*I* + *am*).

If clitics and contractions are not automatically expanded by Stanza, we use a rules-based analysis of orthography to detect some of these common forms and manually expand them. This functionality is currently supported for detection of subject contractions in French and Italian (i.e. *t'aime* to *te* + *aime*), prepositional contractions (i.e. *jusqu'ici* to *jusque* + *ici*), and be-contractions in English (i.e. *you're* to *you* + *are*).

The third type of MWT not typically expanded by Stanza, but which our pipeline uses a lexicon to detect and expand, are single-unit, multi-word forms which are usually joined by an underscore in the CHAT transcription format (MacWhinney 2014) because they are a single semantic form and multiple syntactic words. For instance, the form *pirates_des_Caraïbes* (Pirates of the Caribbean) is one such form, broken into *pirates des Caraïbes*.

We implement this correction functionality as a custom step in the Stanza analysis pipeline; this step takes the "draft" tokenizations from Stanza as input and returns the correct tokenization and word expansions to downstream analysis functions in Stanza—ensuring that POS, GFs, and GRs will be analyzed on the corrected word.

Additionally, the neural tokenizer in Stanza would occasionally mark forms as MWTs when they are simply single-token single-word forms with a punctuation within (i.e. the French word *aujourd'hui*); in those cases, we perform the opposite correction forcing Stanza to treat the resulting token a single word instead of an MWT. These cases are identified and corrected using a lexicon as well.

In final output into the CHAT transcription format, we follow the convention set forth by the CLAN MOR/MEGRASP system (MacWhinney et al., 2012) and join the morphology analyses of multi-word tokens together with a tilde (~), maintaining token-level alignment between the transcript and analysis yet being able to encode multiple words within a token.

**Morphology and Dependency Analysis**

After tokenization and MWT correction, we make no further adjustments to the Stanza morphology, dependency, and feature analysis of each language and simply run the remaining Stanza analysis pipeline with the corrected tokens. Because most Stanza models are trained via the Universal Dependencies dataset, some datasets, such as UD Dutch Alpino (Bouma et al., 2001), will be rich in annotated feature information whereas some others, such as UD Japanese GSD (Nivre et al., 2020), will have little to no GFs annotated. For Japanese, this is true in part because many of the GRs are expressed in separate morphology. Our UD analysis, therefore, carries the design choices of analysis made within these gold datasets. Once this information on POS, GFs, and GRs has been annotated by the Stanza system, we proceed to perform morphology-dependent extraction and correction of the resulting features as a final processing step.

**Morphosyntactic Transcription and Feature Correction**

After analysis by Stanza, we output the extracted GFs using an annotation format very similar to the one used in the MOR/MEGRASP system (described further in https://talkbank.org/manuals/mor.pdf) for the *%mor* and *%gra* lines in CHAT. Our overarching goal is to report the *maximal set* of GFs which 1) can be reported for each language and 2) provide additional information beyond the "default" case.

In accord with these principles, the GFs for aspect, mood, tense, polarity, clusivity, case, type, degree, conjugation (form), and politeness are reported exactly as in the UD annotation specifications. Gender is reported for all tagged genders except "common neutral" (ComNeut); and number is reported for all except singular. For person-hood, fourth and zeroth person are both reported as "fourth person". As in MOR, GFs are joined after the lemma by using a dash "-" and contractions and clitics are marked with ~, as in the earlier MOR standard.

**Dependency Structure**

In addition to creating a %mor line with its analysis of POS and GFs, Batchalign2 also produces a %gra line that encodes the GRs for each utterance. The creation of this GR analysis is the primary goal of the Universal Dependencies project. The encoding involves a directed acyclic graph in which words are connected through unidirectional arcs from the dependent word to its head. Each arc is labeled with a grammatical relation tag taken from the list summarized earlier. Using the GraphViz web service ([https://github.com/xflr6/graphviz](https://github.com/xflr6/graphviz)), one can double-click on a %gra line to produce a display such as the screenshot in Figure 1 which comes from a parental utterance in the Brown/Eve/020000b.cha file on line 44.

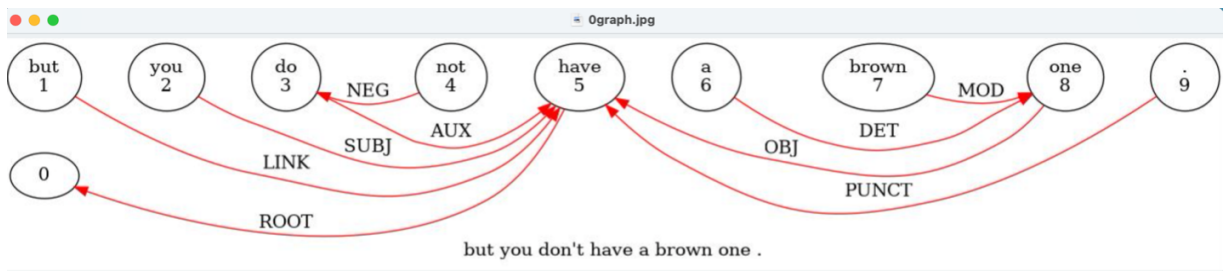

**Figure 1.** *Dependency analysis by UD for an example utterance.*

This graph derives from processing of this utterance:
*MOT:  but you don't have a brown one .
%mor:  cconj|but pron|you-Prs-Nom-S2 aux|do-Fin-Ind-Pres-S2~part|not
       verb|have-Inf-S det|a-Ind-Art adj|brown-Pos-S1 noun|one .
%gra:  1|5|CC 2|5|NSUBJ 3|5|AUX 4|5|ADVMOD 5|8|ROOT 6|8|DET 7|8|AMOD 8|5|OBJ
       9|5|PUNCT

In the %gra line, each word has two numbers and a GR. The first number is its serial position in the utterance and the second is the position of the word to which it is linked through a GR. After the two numbers comes the label on the GR. In Figure 1, for example, we see that the word *one* links to the verb *have* through the OBJ relation, that the word *brown* links to *one* through the MOD relation, and so on. This form of display is essentially the same as what was produced by MEGRASP (Figure 2), although the labels on the arcs are changed and in UD the word *not* is linked to the auxiliary *do* rather than directly to the verb.

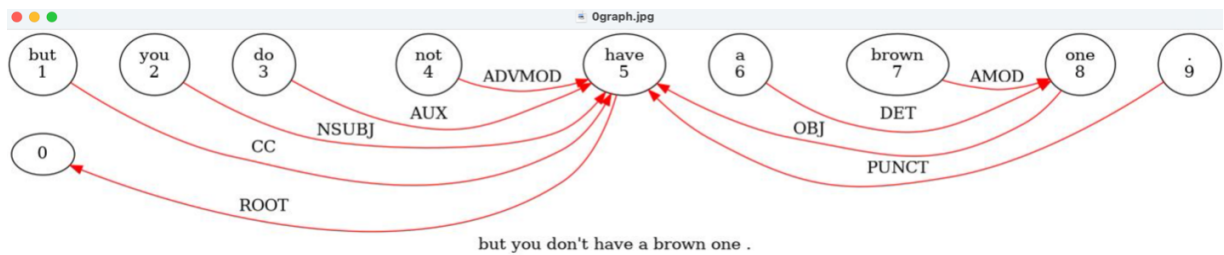

**Figure 2.** *Dependency analysis by MEGRASP for an example utterance.*

### Processing based on UD Analysis

Having tagged corpora in 27 of the languages in CHILDES for POS, GFs, and GRs, we are able to apply many of the TalkBank analytic tools that were earlier available only for English. This opportunity can go a long way toward reducing the WEIRD emphasis in child language studies. Most of these tools and frameworks will work directly, but some require further configuration. We can now use them to compute indices and profiles for the three data formats discussed earlier: longitudinal case studies, cross-sectional group studies, and crosslinguistic comparisons. In other words, having this morphosyntactic information available for all 27 languages benefits not only cross-linguistic comparison, but also the language-internal examination of development for individual children and clinically-important comparison groups within each language. The tools that are available now or which will soon be available include:

1. Basic analysis commands: Researchers could make use of the 26 basic analysis commands in CLAN on all languages prior to running of Batchalign2. However, because most of the languages previously had no %mor or %gra line, analyses were limited to the main speech tier. Now these same programs can run on these additional lines, making many additional types of analyses possible.

2. KIDEVAL: This command combines 57 CLAN analyses into a single package. It includes tracking of the most common GFs in each language, repetitions, vocabulary diversity, error types, MLU (mean length of utterance), and other indicators. In a single command, KIDEVAL can be run on a single transcript or a whole folder of transcripts. It gives both the results for each child on each measure as well as a z-score for the extent to which the child matches a larger comparison group for that measure. The comparison group can be selected for age group in 6-month intervals, participant type (TD, DLD, ASD, etc.) and recording type (narrative, free play, elicited). For this comparison to be meaningful, KIDEVAL needs a comparison sample of at least 25 cases. This is currently possible for Dutch, English, French, Japanese, Mandarin, and Spanish. Construction of comparison corpora for other languages that have sufficient comparison data is in progress.

3. DSS: DSS (Developmental Sentence Score) (Lee, 1974) is a profiling method that focuses on early learning of grammatical morphology and basic syntax in English. Given the new availability of a consistent set of POS, GF and GR tags, it will now be much easier to configure versions of DSS for additional languages.

4. IPSyn: IPSyn (the Index of Productive Syntax) (Scarborough, 1990) is similar to DSS. However, it includes measures of more advanced syntactic structures. Building on recent analyses (MacWhinney et al., 2020; Yang et al., 2021) we can create streamlined, automatic versions of IPSyn for multiple languages.

5. Vocabulary diversity: CLAN provides four measures of vocabulary diversity: TTR (type token ratio), NDW (number of different words), MATTR (moving average type token ratio) (Covington & McFall, 2010), and vocD (Malvern & Richards, 1997). Analysis through MATTR and vocD requires use of lemmas on the %mor line which is now possible across the 27 languages to which UD has been applied.

6. GF analysis: Although a basic level of GF analysis is built into KIDEVAL, there are many types of crosslinguistic analysis that will be best conducted using programs like FREQ on the %mor line across languages. For example, we can now look consistently at learning of tense marking across all these languages and observe how that feature is acquired in comparison with other features.

7. GR analysis: It is now possible to use GraphViz to visualize the syntactic structure for all 27 languages. In addition, Section 7.9.14 of the CLAN manual describes how to use FREQ with the UD %gra line to study the emergence of more complex relations, such as xcomp (a clausal complement without its own subject) or expl:pass (a reflexive marker of a middle or passive clause), as well as combinations of GRs.

8. Cross-tier analysis: We are currently building a new program called FLUPOS for tracking features across multiple coding tiers, including the main line, %mor, %gra, and the %pho line for phonology. One particularly important application of FLUPOS will be to determine the degree to which disfluencies are proportionally higher with certain lexical, morphological, phonological, and syntactic configurations.

The combination of these new %mor and %gra tiers for these 27 languages, along with current analytic methods and ones we plan to build will provide us with a strong quantitative foundation for crosslinguistic analysis of language development. We will be able to track the impact of language structure and input on the development of lexicon, morphology, and syntax in a set of languages that goes well beyond the limits of data from only WEIRD participants.

## Authorship and Contributorship Statement

Houjun Liu is the author of the Batchalign2 program and Brian MacWhinney applied the program to analyze corpora in 27 languages. Both authors shared in the conception of the work, writing of this report, and approval of the final version. Both ensure that questions related to the accuracy or integrity of any part of the work will be appropriately investigated and resolved.

### Data, code and materials availability statement

Batchalign2 is available from https://github.com/talkbank/batchalign2. The utterance segmentation models trained using the manner described in (Liu et al. 2023), as well as fine-tuned, language specific Whisper models, are additionally available for US English and Mandarin; these models are available at https://huggingface.co/talkbank. The tagged corpora for the 27 languages discussed here are available from https://childes.talkbank.org. All the corpora in CHILDES are open-access and have associated DOIs.

### License